%% file: main.tex
\documentclass{article}

\input{preamble}

\input{acronyms}

\begin{document}

\maketitle

\vspace{-0.2in}
\input{sections/abstract}

\input{sections/introduction}

\input{sections/results}

\input{sections/conclusion}

\clearpage

\bibliographystyle{apalike}
\input{sections/bibliography}

\end{document}

%% file: preamble.tex
     \usepackage[final]{neurips_2020}

\usepackage[utf8]{inputenc} %
\usepackage[T1]{fontenc}    %
\usepackage[hidelinks]{hyperref}       %
\usepackage{url}            %
\usepackage{booktabs}       %
\usepackage{amsfonts}       %
\usepackage{nicefrac}       %
\usepackage{microtype}      %
\usepackage{rotating,tabularx}
\usepackage{graphicx}
\usepackage{multirow}
\usepackage{amsmath}
\usepackage{amssymb}
\usepackage{cleveref}
\usepackage{sidecap}
\title{Evaluating the fairness of fine-tuning strategies in self-supervised learning}

\newcommand*\samethanks[1][\value{footnote}]{\footnotemark[#1]}

\author{
  Jason Ramapuram\thanks{Equal contribution. Order determined by \texttt{np.random.uniform}.}, \ \ \ Dan
  Busbridge\samethanks, \ \ \ Russ Webb \\
  Apple \\
\texttt{\{jramapuram,\ dbusbridge,\ rwebb\}@apple.com} \\
}

%% file: acronyms.tex
\usepackage[acronym,smallcaps,nowarn,section,nonumberlist]{glossaries}
\glsdisablehyper
\makeglossaries

\setacronymstyle{long-short}

\newacronym{bn}{BN}{Batch Normalization}
\newacronym{ssl}{SSL}{Self-Supervised Learning}
\newacronym{simclr}{SimCLR}{A Simple framework for Contrastive Learning of visual Representations}

%% file: sections/abstract.tex
\vspace{-0.1in}
\begin{abstract}
  \vspace{-0.1in}
  In this work we examine how fine-tuning impacts the fairness of contrastive \gls{ssl} models. Our findings
indicate that \gls{bn} statistics play a crucial role, and that
updating \emph{only} the \gls{bn} statistics
of a pre-trained \gls{ssl} backbone improves its downstream fairness
(\emph{36\%} worst subgroup, \emph{25\%} mean subgroup gap).
This procedure is competitive with supervised
learning, while taking \emph{4.4$\times$} less time to train and requiring only $0.35\%$ as many parameters to be updated. Finally, inspired by recent work in supervised learning, we find that updating \gls{bn} statistics and training residual skip connections (\emph{12.3\%} of the parameters) achieves parity with a fully fine-tuned model, while taking \emph{1.33$\times$} less time to train.

\end{abstract}

%% file: sections/introduction.tex
\vspace{-0.2in}
\section{Introduction}
\label{sec:introduction}
\vspace{-0.1in}

\gls{ssl} is an effective pre-training strategy in the image
\citep{
DBLP:conf/icml/ChenK0H20,
DBLP:conf/nips/GrillSATRBDPGAP20,
DBLP:journals/corr/abs-2104-14294,
DBLP:conf/nips/CaronMMGBJ20,
DBLP:journals/corr/abs-2103-03230,
DBLP:journals/corr/abs-2105-04906},
language \citep{DBLP:conf/naacl/DevlinCLT19}, video
\citep{DBLP:conf/nips/AlayracRSARFSDZ20} and audio
\citep{DBLP:conf/cvpr/DengDSLL009} domains.
These large scale \gls{ssl} models are
trained without the use of (potentially) biased human annotations, and attain
better than supervised performance when fine-tuned on small sample supervised datasets.
The performance guarantees for many of these large scale \gls{ssl} models is strongly
coupled with the use of \gls{bn}
\citep{DBLP:conf/icml/ChenK0H20,DBLP:journals/corr/abs-2103-03230,DBLP:conf/nips/CaronMMGBJ20,DBLP:conf/nips/AlayracRSARFSDZ20,Fetterman_Albrecht_2020}.
\gls{bn} \citep{DBLP:conf/icml/IoffeS15} tends to favor subgroups of the dataset which
contain more samples, negatively impacting downstream model performance for under-represented subpopulations.

To understand how \gls{ssl} model fairness is impacted by fine-tuning, we evaluate a number of tuning strategies.
We find that the treatment of \gls{bn} statistics is a dominant factor for  determining downstream fairness.
When tuning a linear task head,
freezing \gls{bn} statistics and backbone parameters reduces performance by up to \emph{36\%} in
the worst subgroup fairness metric, whereas allowing  \gls{bn} statistics to update reduces the
performance gap against a fully fine-tuned model, while taking \emph{4.4$\times$}
less time to train and updating only \emph{0.35\%} of the total model parameters.

%% file: sections/results.tex
\vspace{-0.15in}
\section{Results}
\label{sec:results}
\vspace{-0.1in}

Our baseline \gls{ssl} model uses the SimCLR
framework and optimization procedure
\citep{DBLP:conf/icml/ChenK0H20,DBLP:journals/corr/GoyalDGNWKTJH17}, pre-trained
(no labels) on the Celeb-A train split (162,770 samples).
We then attach a linear head to the backbone and evaluate five
scenarios inspired by analysis in supervised learning \citep{DBLP:journals/corr/abs-2003-00152}: fully fine-tuned (\emph{Full FT}); frozen backbone, updating
residual skip connections and \gls{bn} stats (\emph{\gls{bn}
Stats+Skip}); frozen backbone, updating \gls{bn} affine parameters and
\gls{bn} stats (\emph{\gls{bn} Stats+Affine}); frozen backbone, updating
\gls{bn} stats (\emph{\gls{bn} Stats}); and fully frozen backbone
(\emph{Frozen}). Updates are done using supervised information from the Celeb-A
train split.

\input{tables/table_batchnorm}
\input{figures/figure_batchnorm}

We evaluate Celeb-A test split (19,962 samples) using the 40-dimensional binary attribute prediction task. We baseline our \gls{ssl} model
against a strong supervised learning model, which uses the same ResNet50
\citep{DBLP:conf/cvpr/HeZRS16} backbone. To choose
hyper-parameters, we perform a random search (twenty trials) across optimizers
\citep{DBLP:conf/aaai/HuoGH21,DBLP:journals/corr/KingmaB14}, learning rates and schedulers
\citep{DBLP:journals/corr/GoyalDGNWKTJH17,DBLP:journals/corr/abs-1708-07120},
weight decay, training epochs, and linear warmup intervals.
Equivalent compute
budget is used for the SSL fine-tuning and supervised models, and we provide
results for the best performing model from each search.

Quantifying fairness is challenging due to its multifaceted nature \citep{DBLP:journals/corr/abs-2001-07864}, with some facets mutually incompatible \citep{DBLP:journals/corr/FriedlerSV16}.
In this work, taking the $F_1$ score as a performance measure, a fair model maximizes the $F_1$ for the worst treated subgroup ($F_1^{\textrm{worst}}$) and minimizes performance differences across subgroups ($F_1^{\textrm{gap}}$).
Concretely, let
$\mathcal C=\{\textrm{bald},\ldots\}$
be the set of all 40 Celeb-A categories,
$F_1(t | c)$ denote the $F_1$ score achieved on task $t\in\mathcal C$ for the subpopulation with $c\in\mathcal C$ as true\footnote{For example:
$F_1(\textrm{wearing hat }|\textrm{ blurry})$
is the $F_1$ score for \emph{blurry} images when predicting \emph{wearing hat}.
}, and $F_1(t | \neg c)$ equivalently for populations with $c$ as false.
We define
\begin{align}
	F_1^{\textrm{gap}}(t,c)
	&=
	\left|
	\left(F_1(t | c) - F_1(t | \neg c)\right)
	\right|,
	&
	F_1^{\textrm{worst}}(t, c)
	&=
	\min \left(F_1(t | c),F_1(t | \neg c)\right)
	.
\end{align}
Model performance across the 1560 $(t, c)$ combinations\footnote{We omit on-diagonal ($t=c$) terms to ensure all metric components are well-defined.} is summarized in \Cref{fig:batch-norm} and \Cref{table:batch-norm-comparison}.

%% file: tables/table_batchnorm.tex
\begin{table}[ht]
\small
\centering
\scalebox{0.85}{\hspace{-1.4cm}\parbox{1.0\linewidth}{
\centering
\begin{tabular}{llllllllllllll}
\toprule
                                           &    Training Procedure & \rotatebox{90}{bald} & \rotatebox{90}{double chin} & \rotatebox{90}{chubby} & \rotatebox{90}{wearing necktie} & \rotatebox{90}{wearing necklace} & \rotatebox{90}{no beard} & \rotatebox{90}{straight hair} & \rotatebox{90}{big lips} & \rotatebox{90}{wavy hair} & \rotatebox{90}{male} & \rotatebox{90}{wearing lipstick} & \rotatebox{90}{all} \\
\midrule
                                 $\rho(c)$ &                       &                  .02 &                         .05 &                    .06 &                             .07 &                              .12 &                      .17 &                           .21 &                      .24 &                       .32 &                  .42 &                              .47 &                     \\
\midrule \multirow{6}{*}{Gap $\downarrow$} &          SSL (Frozen) &                $.19$ &                       $.03$ &                  $.04$ &                           $.11$ &                            $.03$ &                    $.08$ &                $\mathbf{.03}$ &           $\mathbf{.04}$ &                     $.03$ &                $.05$ &                   $\mathbf{.00}$ &               $.04$ \\
                                           &        SSL (BN Stats) &                $.17$ &              $\mathbf{.00}$ &                  $.02$ &                           $.08$ &                            $.02$ &                    $.04$ &                         $.03$ &                    $.04$ &                     $.02$ &                $.01$ &                            $.02$ &               $.04$ \\
                                           & SSL (BN Stats+Affine) &                $.16$ &                       $.01$ &                  $.01$ &                           $.09$ &                   $\mathbf{.02}$ &                    $.04$ &                         $.04$ &                    $.04$ &                     $.02$ &                $.02$ &                            $.01$ &               $.04$ \\
                                           &   SSL (BN Stats+Skip) &                $.14$ &                       $.01$ &                  $.01$ &                           $.09$ &                            $.03$ &                    $.03$ &                         $.03$ &                    $.04$ &            $\mathbf{.01}$ &                $.01$ &                            $.03$ &               $.04$ \\
                                           &         SSL (Full FT) &       $\mathbf{.10}$ &                       $.01$ &         $\mathbf{.01}$ &                           $.08$ &                            $.02$ &                    $.03$ &                         $.03$ &                    $.04$ &                     $.01$ &                $.00$ &                            $.02$ &      $\mathbf{.03}$ \\
                                           &            Supervised &                $.12$ &                       $.01$ &                  $.01$ &                  $\mathbf{.08}$ &                            $.02$ &           $\mathbf{.03}$ &                         $.04$ &                    $.04$ &                     $.01$ &       $\mathbf{.00}$ &                            $.03$ &               $.03$ \\
\midrule \multirow{6}{*}{Worst $\uparrow$} &          SSL (Frozen) &                $.30$ &                       $.45$ &                  $.45$ &                           $.38$ &                            $.46$ &                    $.37$ &                         $.46$ &                    $.47$ &                     $.45$ &                $.37$ &                            $.40$ &               $.43$ \\
                                           &        SSL (BN Stats) &                $.56$ &                       $.72$ &                  $.70$ &                           $.64$ &                            $.71$ &                    $.64$ &                         $.70$ &                    $.72$ &                     $.70$ &                $.62$ &                            $.63$ &               $.69$ \\
                                           & SSL (BN Stats+Affine) &                $.58$ &                       $.73$ &                  $.72$ &                           $.65$ &                            $.72$ &                    $.65$ &                         $.71$ &                    $.73$ &                     $.72$ &                $.64$ &                            $.66$ &               $.70$ \\
                                           &   SSL (BN Stats+Skip) &                $.60$ &                       $.74$ &                  $.73$ &                           $.66$ &                            $.72$ &                    $.67$ &                         $.72$ &                    $.74$ &                     $.73$ &                $.64$ &                            $.65$ &               $.71$ \\
                                           &         SSL (Full FT) &       $\mathbf{.65}$ &              $\mathbf{.75}$ &         $\mathbf{.74}$ &                           $.67$ &                   $\mathbf{.73}$ &           $\mathbf{.67}$ &                $\mathbf{.73}$ &           $\mathbf{.74}$ &            $\mathbf{.74}$ &       $\mathbf{.65}$ &                   $\mathbf{.66}$ &      $\mathbf{.72}$ \\
                                           &            Supervised &                $.63$ &                       $.74$ &                  $.73$ &                  $\mathbf{.67}$ &                            $.73$ &                    $.67$ &                         $.71$ &                    $.74$ &                     $.73$ &                $.65$ &                            $.64$ &               $.71$ \\
\bottomrule
\end{tabular}
}}
\caption{$c-$wise \emph{gap} and \emph{worst} scores: $F_1^{\textrm{(gap | worst)}}(c)=N_c^{-1}\sum_{t\neq c}F_1^{(\textrm{gap | worst)}}(t,c)$ with under-representation statistic: $\rho(c)=\tfrac{\min\left(N(c),N(\neg c)\right)}{N(c)+N(\neg c)}$. Attributes presented are uniformly distributed across $\rho(c)$, displaying balanced $(\rho\sim .5)$ and imbalanced $(\rho\ll 0.5)$ model behaviour. We note that the \emph{gap} (\emph{worst}) statistic is smaller (larger/worse) for large $\rho$. \emph{All} = $\textrm{median}_c[F_1^{\textrm{(gap | worst)}}(c)]$.}
\label{table:batch-norm-comparison}
\end{table}

%% file: figures/figure_batchnorm.tex
\sidecaptionvpos{figure}{c}
\begin{SCfigure}
  \includegraphics[width=0.49\textwidth]{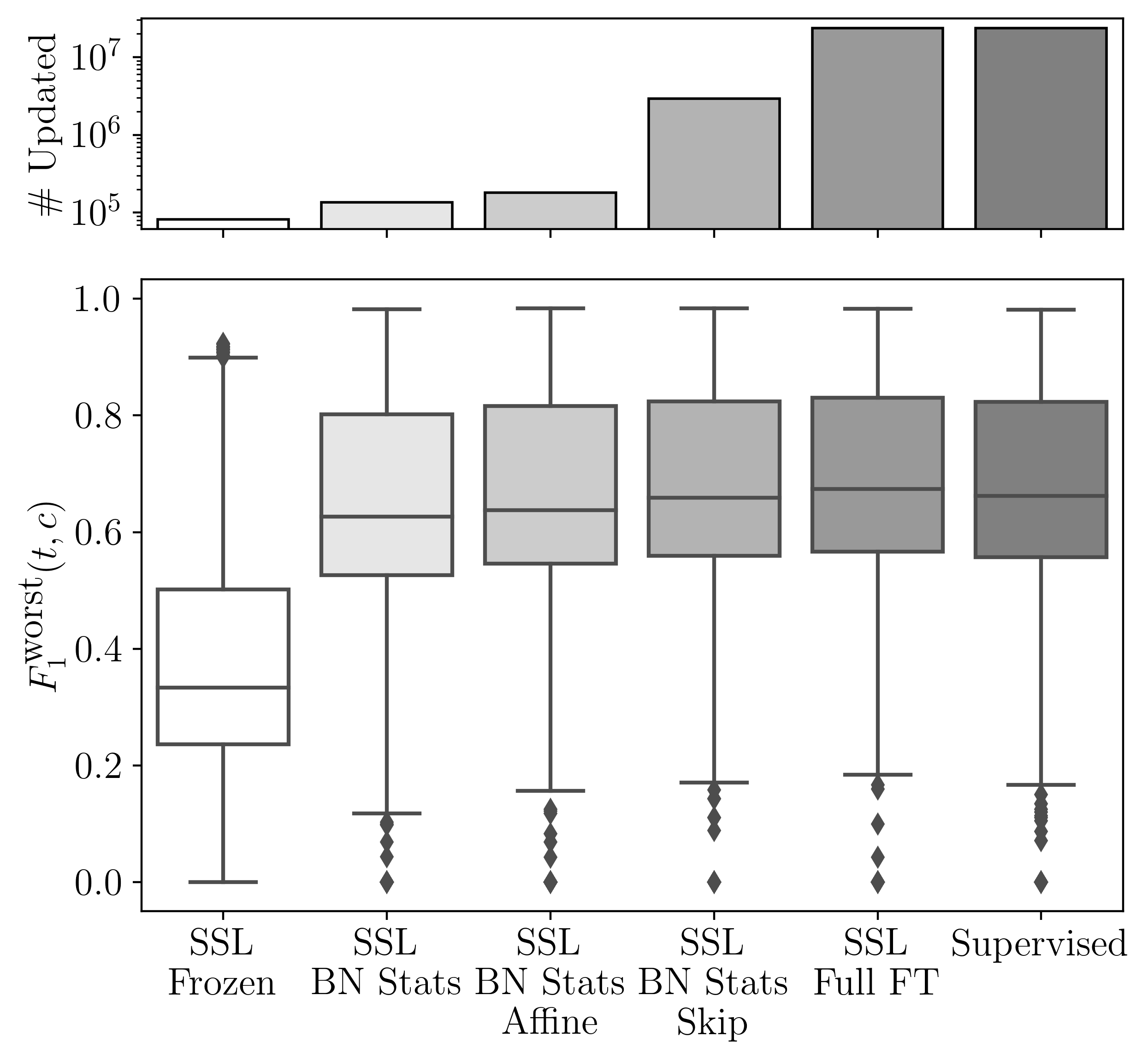}
  \caption{\emph{Top}: Total number of parameters and buffers updated per model. 
  \emph{Bottom}: Distribution of $F_1^{\textrm{worst}}$ over 1560 $(t, c)$ combinations on the Celeb-A test split. Individual attribute thresholds are calibrated on the Celeb-A train split with no $c-$conditioning.
  Higher $F_1^{\textrm{worst}}$ distribution indicates a fairer model in the absolute sense.
  \emph{SSL (Full FT)}, \emph{SSL (\gls{bn} Stats+Skip)} and \emph{Supervised} perform similarly across the board, with the fully fine-tuned SSL model being marginally better. \emph{SSL (frozen)} drastically underperforms compared to these four models, however, the performance gap can be closed by simply allowing the running statistics to be updated, yielding \emph{SSL (\gls{bn} stats)}.}
  \label{fig:batch-norm}
\end{SCfigure}

%% file: sections/conclusion.tex
\vspace{-0.1in}
\section{Conclusion}
\label{sec:conclusion}
\vspace{-0.05in}

Models that produce fair representation vectors can directly improve the fairness
of any downstream task that uses them.
These models
have the ability to affect fairness at a large scale, through the use of
developer APIs.
In this work, we quantify the the effect that various
fine-tuning strategies play in downstream fairness, and observe the crucial role
played by \gls{bn} statistics.
We demonstrate that \emph{only} updating \gls{bn} statistics
minimizes the gap between an end-to-end trained model and a frozen SSL model,
improving worst case subgroup fairness by \emph{36\%} and taking
\emph{4.4$\times$} less time to train.

%% file: sections/bibliography.tex
\bibliography{libraries/selfsup,libraries/celeba,libraries/models,libraries/other}